# Value-Directed Sampling Methods for Monitoring POMDPs

**Pascal Poupart**
Department of Computer Science
University of Toronto
Toronto, ON M5S 3H5
*ppoupart@cs.toronto.edu*

**Luis E. Ortiz**
Computer Science Department
Brown University
Providence, RI, USA 02912-1210
*leo@cs.brown.edu*

**Craig Boutilier**
Department of Computer Science
University of Toronto
Toronto, ON M5S 3H5
*cebly@cs.toronto.edu*

## Abstract

We consider the problem of approximate belief-state monitoring using particle filtering for the purposes of implementing a policy for a partially observable Markov decision process (POMDP). While particle filtering has become a widely used tool in AI for monitoring dynamical systems, rather scant attention has been paid to their use in the context of decision making. Assuming the existence of a value function, we derive error bounds on decision quality associated with filtering using importance sampling. We also describe an adaptive procedure that can be used to dynamically determine the number of samples required to meet specific error bounds. Empirical evidence is offered supporting this technique as a profitable means of directing sampling effort where it is needed to distinguish policies.

## 1 Introduction

Considerable attention has been devoted to partially observable Markov decision processes (POMDPs) [19] as a model for decision-theoretic planning. Their generality allows one to seamlessly model sensor and action uncertainty, uncertainty in the state of knowledge, and multiple objectives [1, 4]. Despite their attractiveness as a conceptual model, POMDPs are intractable and have found practical applicability in only limited special cases.

The predominant approach to the solution of POMDPs involves generating an optimal or approximate *value function* via dynamic programming: this value function maps *belief states* (or distributions over system states) into optimal expected value, and implicitly into an optimal choice of action. Constructing such value functions is computationally intractable and much effort has been devoted to developing approximation methods or algorithms that exploit specific problem structure. Potentially more troublesome is the problem of *belief state monitoring*—maintaining a belief state over time as actions and observations occur so that the optimal action choice can be made. This too is generally intractable, since a distribution must be maintained over the set of system states, which has size exponential in the number of system variables. While value function construction is an offline problem, belief state monitoring must be effected in real time, hence its computational demands

are considerably more pressing.[1]

One important family of approximate belief state monitoring methods is the *particle filtering* or *sequential Monte Carlo* approach [6, 13]. A belief state is represented by a random sample of system states, drawn from the true state distribution. This set of *particles* is propagated through the system dynamics and observation models to reflect the system evolution. Such methods have proven quite effective, and have been applied in many areas of AI such as vision [11] and robotics [21].

While playing a large role in AI, the application of particle filters to decision processes has been limited. While Thrun [20] and McAllester and Singh [14] have considered the use of sampling methods to solve POMDPs, we are unaware of studies using particle filters in the implementation of a POMDP policy. In this paper we examine just this, focusing on the use of fairly standard importance sampling techniques. Assuming a POMDP has been solved (i.e., a value function constructed), we derive bounds on the error in decision quality associated with particle filtering with a given number of samples. These bounds can be used *a priori* to determine an appropriate sample size, as well as forming the basis of a *post hoc* error analysis. We also devise an adaptive scheme for dynamic determination of sample size based on the probability of making an (approximately) optimal action choice given the current set of samples at any stage of the process. We note that similar notions have been applied to the problem of influence diagram evaluation by Ortiz and Kaelbling [15] with good results—our approach draws much from this work, though with an emphasis on the sequential nature of the decision problem.

A key motivation for taking a value-directed approach to sampling lies in the fact that monitoring is an online process that must be effected quickly. One might argue that if the state space of a POMDP is large enough to require sampling for monitoring, then its state space is too large to hope to solve the POMDP. To counter this claim, we note first that recent algorithms [2, 9] based on factored representations, such as dynamic Bayes nets (DBNs), can often solve POMDPs without explicit state space enumeration and produce reasonably compact value function representations. Unfortunately, such representations do not generally translate into effective (exact) belief monitoring schemes [3]. Even in cases where a POMDP must be solved in a traditional "flat" fashion, we typically have the luxury of compiling a value function offline. Thus, even for large POMDPs, we might reasonably expect to have value function information (either exact or approximate) available to direct the monitoring process. The fact that one is able to produce a value function *offline* does not imply the ability to monitor the process exactly in a timely *online* fashion.

---

[1]While techniques exist for generating finite-state controllers for POMDPs, there are still reasons for wanting to use value-function-based approaches [17].

We overview POMDPs, structured solution techniques, and monitoring in Section 2. Section 3 describes a basic particle filtering scheme for POMDPs and analyzes its error. We also describe a dynamic sample generation scheme that relies on ideas from group sequential sampling. We examine this model empirically in Section 4, and conclude with a discussion of future directions.

## 2 POMDPs and Belief State Monitoring

### 2.1 Solving POMDPs

A partially observable Markov decision process (POMDP) is a general model for decision making under uncertainty. Formally, we require the following components: a finite state space $\mathcal{S}$; a finite action space $\mathcal{A}$; a finite observation space $\mathcal{Z}$; a transition function $T : \mathcal{S} \times \mathcal{A} \rightarrow \Delta(\mathcal{S})$;[2] an observation function $O : \mathcal{S} \times \mathcal{A} \rightarrow \Delta(\mathcal{Z})$; and a reward function $R : \mathcal{S} \rightarrow \mathbf{R}$. Intuitively, the transition function $T(s, a)$ determines a distribution over next states when an agent takes action $a$ in state $s$. This captures uncertainty in action effects. The observation function reflects the fact that an agent cannot generally determine the true system state with certainty (e.g., due to sensor noise). Finally $R(s)$ denotes the immediate reward associated with $s$.

The rewards obtained over time by an agent adopting a specific course of action can be viewed as random variables $R^{(t)}$. Our aim is to construct a *policy* that maximizes the expected sum of discounted rewards $E(\sum_{t=0}^{\infty} \beta^t R^{(t)})$ (where $\beta$ is a discount factor less than one). It is well-known that an optimal course of action can be determined by considering the fully observable *belief state MDP*, where *belief states* (distributions over $\mathcal{S}$) form states, and a policy $\pi : \Delta(\mathcal{S}) \rightarrow \mathcal{A}$ maps belief states into action choices. In principle, dynamic programming algorithms for MDPs can be used to solve this problem. A key result of Sondik [19] showed that the value function $V$ for a finite-horizon problem is piecewise-linear and convex and can be represented as a finite collection of $\alpha$-vectors.[3] Specifically, one can generate a collection $\aleph$ of $\alpha$-vectors, each of dimension $|\mathcal{S}|$, such that $V(b) = \max_{\alpha \in \aleph} b\alpha$. Figure 1 illustrates a collection of $\alpha$-vectors with the upper surface corresponding to $V$. We define $ma(b) = \arg \max_{\alpha \in \aleph} b\alpha$ to be the maximizing $\alpha$-vector for belief state $b$.

Each $\alpha \in \aleph$ corresponds to the expected value of executing an implicit *conditional plan* at a given belief state. This conditional plan, $\pi(\alpha)$, has the form $\langle a; o_1, \pi_1; o_2, \pi_2; \cdots o_n, \pi_n \rangle$, where $a$ is an action, $o_i$ is an observation, and $\pi_i$ is itself a conditional plan. Intuitively, a plan of this form denotes the performance of action $a$ followed by execution of the remaining plan $\pi_i$ in response to observation $o_i$. We denote by $A(\alpha)$ the (first) action $a$ of $\pi(\alpha)$. Given belief state $b$, the agent should execute the action with the maximizing $\alpha$-vector: $A(ma(b))$. Indeed, if one has access to the entire plan $\pi(ma(b))$, this plan should be executed to termination. We note, however, that the plans $\pi(\alpha)$ are rarely recorded explicitly.

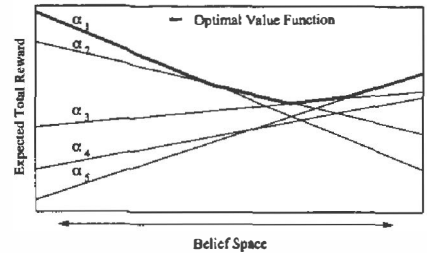

Figure 1: Geometric View of Value Function

One difficulty with these classical approaches is the fact that the $\alpha$-vectors may be difficult to manipulate. A system characterized by $n$ random variables has a state space size that is exponential in $n$. Thus manipulating a single $\alpha$-vector may be intractable for complex systems.[4] Fortunately, it is often the case that an MDP or POMDP can be specified very compactly by exploiting structure (such as conditional independence among variables) in the system dynamics and reward function [1]. Representations such as dynamic Bayes nets (DBNs) can be used, and schemes have been proposed whereby the $\alpha$-vectors are computed directly in a factored form by exploiting this representation.

Boutilier and Poole [2], for example, represent $\alpha$-vectors as decision trees in implementing Monahan's algorithm. Hansen and Feng [9] use algebraic decision diagrams (ADDs) as their representation in their version of incremental pruning. The empirical results in [9] suggest that such methods can make reasonably sized problems solvable. Furthermore, factored representations will likely facilitate good approximation schemes.

### 2.2 Belief State Monitoring

Given a value function represented using a collection $\aleph$ of $\alpha$-vectors, implementation of an optimal policy requires that one maintain a belief state over time in order to apply it to $\aleph$. Given belief state $b^t$ at time $t$, we determine $a^t = A(ma(b^t))$, execute $a^t$, make a subsequent observation $o^{t+1}$, then update our belief state to obtain $b^{t+1}$. The process is then repeated. Belief state monitoring is effected by computing $b^{t+1} = \Pr(S|b^t, a^t, o^{t+1})$, which involves straightforward Bayesian updating. We denote by $T(b, a, o)$ the update of any belief state $b$ by action $a$ and observation $o$. We inductively define

$$T(b, a_1, o_1, \cdots, a_n, o_n) =$$
$$\overline{T(T \cdots (T(b, a_1, o_1), \cdots, a_{n-1}, o_{n-1}) a_n, o_n)}$$

Even if the value function can be constructed in a compact way, the *monitoring problem* itself is not generally tractable, since each belief state is a vector of size $|S|$. Unfortunately, even using DBNs does not alleviate the difficulty, since correlations tend to "bleed through" the DBN, rendering most (if not all) variables dependent after a time [3]. Thus compact representation of the exact belief state is typically impossible. Belief state approximation is therefore often required. At any point in time we have an approximation $\tilde{b}^t$ of the true belief state $b^t$, and must make our decisions based on this approximate belief state. While several methods for belief state approximation can be used (including projection, aggregation, and variational methods), and important class of techniques for dynamic problems is *sampling* or *simulation* methods.

## 3  Particle Filtering for POMDPs

In this section we examine the impact of particle filtering on decision quality in POMDPs. We first describe a typical sequential importance sampling algorithm, and discuss the use of partial evidence integration (EI) in the DBN to help keep samples on track. We then analyze the error induced by one stage of belief state approximation and show how partial EI allows this analysis to be carried through multiple stages (in a way that is not possible otherwise).

### 3.1  A Basic Filtering Method for POMDPs

Assume we have been provided with the value function for a specific POMDP $M$. This value function is represented by a finite collection $\aleph$ of $\alpha$-vectors. We assume an infinite-horizon model so that we have a single set $\aleph$. We also assume that $\aleph$ is of a manageable size, and that the vectors themselves are represented compactly (using ADDs, decision trees, linear combinations of basis functions, or some other representation). We emphasize, however, that even if the value function is represented in standard state form, approximate monitoring is often needed. We note that our methods can be applied to approximate value functions, though our analysis assumes an exact set $\aleph$.

Implementation of the policy induced by this value function requires that a belief state $\tilde{b}^t$ be maintained over all times $t$. At any point in time we assume an approximation $\tilde{b}^t$ of the true belief state $b^t$, and make our decisions based on this approximate belief state.

The basic procedure we consider is the use of a particle filter for monitoring, with the approximate belief states so generated used for action selection in the POMDP. At any time $t$, we have a collection $\tilde{b}^t$ of $n^t$ weighted *particles*, or system states, approximating the true distribution $b^t$. Each particle is a pair $\langle s^t_{(i)}, w^t_{(i)} \rangle$. We often simply write $s^t_{(i)}$ to refer to the $i^{th}$ particle ($i \leq n^t$). The total weight of the particle set $\tilde{b}^t$ is $w^t = \sum w^t_{(i)}$. The particle set $\tilde{b}^t$ represents the following distribution (which we also refer to as $\tilde{b}^t$):

$$\tilde{b}^t(s) = \frac{\sum\{w^t_{(i)} : s^t_{(i)} = s\}}{w^t}$$

Given this approximation $\tilde{b}^t$ of $b^t$, action selection will take place in the POMDP as if $\tilde{b}^t$ were the true distribution. Thus, we let $a^t = A(ma(\tilde{b}^t))$, execute action $a^t$, and make observation $o^{t+1}$. Our new approximate belief state $\tilde{b}^{t+1}$ is generated by repeating the following steps until $n^{t+1}$ is greater than some desired threshold:

1. Draw a state $s^t$ from the distribution $\tilde{b}^t$.

2. Draw a state $s^{t+1}$ from the distribution $\Pr(S^{t+1}|s^t, a^t)$.

3. Compute $w = \Pr(o^{t+1}|s^t, a^t, s^{t+1})$

4. Add sample $\langle s^{t+1}_{(i)}, w^{t+1}_{(i)} \rangle = \langle s^{t+1}, w \rangle$ to $\tilde{b}^{t+1}$ and add $w$ to total weight $w^t$.

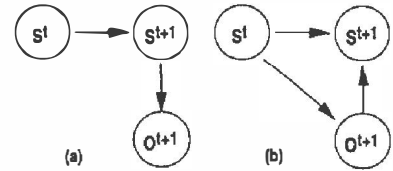

Figure 2: Partial Evidence Integration

This sequential importance sampling procedure induces a consistent, though biased, estimate $\tilde{b}^{t+1}$ of $b^{t+1}$, and will converge to the true distribution according to the usual convergence results. The significance of this method lies in the fact that, for a great many systems, it is easy to sample successor states according to the system dynamics (i.e., sample from the conditional distribution in Step 2), and to evaluate the observation probabilities for given states (i.e., compute the weights in Step 3). In contrast, direct computation of $\Pr(S^{t+1}|b^t, a^t, o^{t+1})$ is generally intractable.

### 3.2  Evidence Integration

One difficulty with the filtering algorithm above is that the samples generated at time $t + 1$ are not influenced by observation $o^{t+1}$, which often allows particles to drift from the true belief state. Since we assume a DBN representation of dynamics, partial *evidence integration (EI)* or arc reversal [8] can be used to partially alleviate this problem [13]. The generic structure of a DBN (assuming a fixed action) is shown in Figure 2(a); reversing the arc from $S^{t+1}$ to $O^{t+1}$ results in a network shown in Figure 2(b). With this structure, given a particle $s^t_{(i)}$ and observation $o^{t+1}$, a particle $s^{t+1}_{(i)}$ can be drawn directly. Of course, the reweighting given $o^{t+1}$ must now be applied to the particles in $\tilde{b}^t$. This gives rise to the following particle filtering procedure used throughout the remainder of the paper:

(a) Given particle set $\tilde{b}^t$, select action $a^t = A(ma(\tilde{b}^t))$, and observe $o^{t+1}$;

(b) Reweight samples $s^t_{(i)}$ according to $\Pr(o^{t+1}|s^t, a^t)$ and normalize to produce $\tilde{b}^{t'}$;

(c) Draw some number of particles $s^t_{(i)}$ according to $\tilde{b}^{t'}$;

(d) Sample particles $s_{(i)}^{t+1}$ given drawn prior particles $s_{(i)}^t$ and $o^{t+1}$ to produce $\tilde{b}^{t+1}$.

Note that the reweighted distribution $\tilde{b}^{t'}$ is an approximation of $\Pr(S^t|a^0,\cdots a^t, o^1, \cdots, o^{t+1})$ in contrast to $\tilde{b}^t$, which represents $\Pr(S^t|a^0, \cdots a^{t-1}, o^1, \cdots, o^t)$.

When the DBN is factored, the arc reversal process can often be fairly expensive, since it increases the connectivity of the network. However, the reversal process can take advantage of the structure in CPTs represented as, say, decision trees or ADD. In this way, the usual exponential increase in table size with the number of added parents is often circumvented [5]. We use structured arc reversal techniques in our experiments.

### 3.3 One-Stage Analysis

As a precursor to bounding the error in decision quality associated with particle filtering, we consider the error induced by one stage of approximation only (and acting using exact inference at all other stages). We first note the following important fact regarding POMDPs:

**Fact 1** *Let $b^t$, $\tilde{b}^t$ be two belief states s.t. $ma(b^t) = ma(\tilde{b}^t)$. For any sequence of $k$ observations and actions, let $b^{t+k} = T(b^t, a^t, o^{t+1}, \cdots a^{t+k-1}, o^{t+k})$ and $\tilde{b}^{t+k} = T(\tilde{b}^t, a^t, o^{t+1}, \cdots a^{t+k-1}, o^{t+k})$. Then $ma(b^{t+k}) = ma(\tilde{b}^{t+k})$.*

This implies that, if we approximate $b^t$ at time $t$ in such a way that $\tilde{b}^t$ has the same maximizing $\alpha$-vector as $b^t$, then we will: (a) choose the correct action at state $t$; and (b) choose the optimal action at all subsequent stages if we monitor the process exactly (w.r.t. $\tilde{b}^t$) at all subsequent stages.

Now, assume we have been able to exactly compute $b^{t-1}$, have selected and executed action $a^{t-1}$ and made observation $o^t$. Furthermore, assume that we can compute $\Pr(S^{t-1}|a^{t-1}, o^t)$ exactly. With these assumptions, we can sample directly from the distribution $b^t = T(b^{t-1}, a^{t-1}, o^t)$ using the (arc-reversed) DBN to obtain an unbiased *estimate* $\tilde{b}^t$ of $b^t$. We analyze the error associated with selecting an $\alpha$-vector that has maximum expected value w.r.t. $\tilde{b}^t$ and executing its conditional plan *to completion* (or equivalently, acting using exact monitoring from that point on).

Let $\{s_{(i)}^t\}$ be a collection of $n^t$ state samples drawn from $b^t$. The value of any $\alpha \in \aleph$ applied to true belief state $b^t$ is:

$$\alpha(b^t) = \alpha \cdot b^t = E_{b^t}[\alpha(s)] = V_\alpha^t$$

where $\alpha(s)$ denotes the value of $\alpha$ at state $s$ (i.e., the $s^{th}$ component of $\alpha$) and $E_{b^t}$ denotes expectation with respect to distribution $b^t$. Thus the value of $\alpha$ can be viewed as a random variable whose expectation (w.r.t. $b^t$) is $V_\alpha^t$. As such, each term $\alpha(s_{(i)}^t)$ is a sample of this random variable and the average of these is an unbiased estimate $\tilde{V}_\alpha^t$ of $V_\alpha^t$. We can apply (one-sided) Hoeffding bounds to determine the accuracy of this estimate. Specifically:

$$\Pr(\tilde{V}_\alpha^t \le V_\alpha^t + \varepsilon) \ge 1 - e^{-2n^t \varepsilon^2/R_\alpha^2}$$
$$\Pr(\tilde{V}_\alpha^t \ge V_\alpha^t - \varepsilon) \ge 1 - e^{-2n^t \varepsilon^2/R_\alpha^2}$$

where $R_\alpha$ is the *range* of values that can be taken by $\alpha$ (i.e., $R_\alpha = \max_S\{\alpha(s)\} - \min_S\{\alpha(s)\}$).

Given a particular confidence threshold $\delta$ and a sample set of size $n^t$ we can produce a (one-sided) error bound $\varepsilon_\alpha$ on the accuracy of our estimate $\tilde{V}_\alpha^t$:

$$\varepsilon_\alpha = \sqrt{\frac{R_\alpha^2 \ln(\frac{1}{\delta})}{2n^t}} \tag{1}$$

The required sample size given error tolerance $\varepsilon$ and confidence threshold $\delta$ for the estimation of $V_\alpha^t$ is:

$$N_\alpha^t(\varepsilon, \delta) = \frac{R_\alpha^2 \ln(\frac{1}{\delta})}{2\varepsilon^2} \tag{2}$$

We can also bound the *simultaneous* confidence that *each* of our estimates of each $\alpha(b^t)$ has (one-sided) precision $\varepsilon$ with probability $1 - \delta$. Decreasing $\delta$ to $\frac{\delta}{|\aleph|}$ in Eq. 2 and maximizing over all $\alpha$, we obtain the sample size $N^t(\varepsilon, \delta)$:

$$N^t(\varepsilon, \delta) = \max_{\alpha \in \aleph} N_\alpha^t\left(\varepsilon, \frac{\delta}{|\aleph|}\right) \tag{3}$$

Choosing the maximizing $\alpha$-vector using an approximate $\tilde{b}^t$ with sample size $N^t(\varepsilon, \delta)$ ensures that a $2\varepsilon$-optimal choice is made with probability at least $1 - \delta$; if the error associated with (arbitrary) nonoptimal behavior is bounded by $h$, then the one-step approximation error is given by the following:

**Theorem 2** *If belief state $\tilde{b}^t$ is approximated with $N^t(\varepsilon, \delta)$ particles, with exact monitoring used at all other stages of the process, then the error $E$ (i.e., difference in expected value of the policy implemented and the optimal policy) is bounded by*

$$E \le \beta^{t+1}(2\varepsilon(1-\delta) + \delta h)$$

Here the error incurred is discounted by $\beta^{t+1}$ to reflect the fact that the approximation error occurs at stage $t$ of the process. Note that the error $h$ on nonoptimal behavior can be easily bounded (rather loosely) using

$$h \le \max_\alpha \max_s \alpha_s - \frac{\beta \min_s\{R(s)\}}{1-\beta}$$

though simple domain analysis will generally yield much tighter bounds on $h$.

One can also perform a *post hoc* analysis on the choice of $\alpha$-vector to determine if an *optimal* choice has been made with high probability. Assuming $n^t$ samples have been generated, let $\varepsilon_\alpha^t$ be the error level determined by Eq. 1 using $n^t$ (this is generally tighter than the $\varepsilon$ used to determine sample size in Eq. 3 since we are looking at a *specific* vector).

**Corollary 3** *Let $\alpha^t = ma(\tilde{b}^t)$ and suppose that*

$$\tilde{b}^t \alpha^t - \varepsilon_{\alpha^t}^t + \tau \ge \tilde{b}^t \alpha + \varepsilon_\alpha^t, \ \forall \alpha \in \aleph \setminus \{\alpha^t\}$$

*Then with probability at least $1 - \delta$ a $\tau$-optimal policy will be executed, and our error is bounded by:*

$$E \le \beta^{t+1}(\tau(1-\delta) + \delta h)$$

The parameter $\tau$ represents the degree to which the value of the second-best $\alpha$-vector may exceed the value of the best at $\tilde{b}^t$ in the worst-case. Note that this relationship must hold for some $\tau \leq 2\varepsilon$. If the relationship holds for $\tau = 0$ (i.e., there is $2\varepsilon$-*separation* between the maximizing vector and all other vectors at belief state $\tilde{b}^t$) then we are executing the optimal policy with probability at least $1 - \delta$ and our error is bounded by $\beta^{t+1}\delta h$.

### 3.4  Multi-stage Analysis

The analysis above assumes that once an $\alpha$-vector is chosen, the plan corresponding to that vector will be implemented over the problem's horizon. In fact, once the first action $A(\alpha)$ is taken, the next action will be dictated by repeating the procedure on the subsequent approximate belief state. Due to further sampling error, the next action chosen may not be the "correct" continuation of the plan $\pi(\alpha)$. Thus we have no assurances that the $2\varepsilon$-optimal policy will be implemented with high probability. In what follows, we assume that our sample size and approximate belief state $\tilde{b}^t$ are such that $\tau = 0$ at every point in time (i.e., our approximate beliefs always give at least $2\varepsilon$-separation for the optimal vector). We discuss this assumption further below.

We make some preliminary observations and definitions before analyzing the accumulated error.

- We first note that $\tilde{b}^{t+1}$ is an unbiased estimate of the distribution $T(\tilde{b}^t, a^t, o^{t+1})$. Though particle filtering does not ensure that $\tilde{b}^{t+1}$ is unbiased with respect to the true belief state $b^{t+1}$, our evidence integration procedure and reweighting scheme produce "locally" unbiased estimates. To see this, notice that the distribution $\tilde{b}^{t'}$ obtained by reweighting $\tilde{b}^t$ w.r.t. $o^{t+1}$ corresponds to *exact* inference assuming the distribution $\tilde{b}^t$ is correct for $S^t$. (This exact computation is tractable precisely because of the sparse nature of this approximate "prior" on $S^t$.) Thus, the procedure for generating samples of $S^{t+1}$ using $\tilde{b}^t$ is a simple forward propagation *without* reweighting, and thus provides an unbiased sample of $T(\tilde{b}^t, a^t, o^{t+1})$.

- Let us say that a *mistake* is made at stage $t$ if $ma(\tilde{b}^{t+1})$ is not optimal w.r.t. $T(\tilde{b}^t, a^t, o^{t+1})$. In other words, due to sampling error, the approximate belief state $\tilde{b}^{t+1}$ differed from the "true" belief state one would have generated using exact inference w.r.t. $\tilde{b}^t$ in such a way as to preclude an optimal policy choice.

We can now analyze the error in decision quality associated with acting under the assumption that $\tau = 0$. Let stage $t$ be the first stage at which a mistake is made. If this is the case, we have that $ma(\tilde{b}^{k+1}) = ma(T(\tilde{b}^k, a^k, o^{k+1}))$ for all $k < t$. By Fact 1, this means that $ma(b^k) = ma(\tilde{b}^k)$ for all $k < t$ (where $b^k$ is the true stage $k$ belief state one would obtain by exact monitoring). Thus, if stage $t$ is the first stage at which a mistake is made, we have acted exactly as we would have using exact monitoring for the first $t$ stages of the process. Since our sampling process produces an unbiased estimate $\tilde{b}^{k+1}$ of $T(\tilde{b}^k, a^k, o^{k+1})$ at each stage, the probability with which no mistake is made before stage $t$ is at least $(1 - \delta)^{t-1}$. Assuming a worst-case bound of $h$ on the performance of an incorrect choice (w.r.t. the optimal policy) at any stage (which is thus independent of any further mistakes being made), we have expected error $E$ on the sampling strategy where $N(\delta, \varepsilon)$ samples are generated at each stage; $E$ is bounded as follows:

**Theorem 4**

$$E \leq \sum_{t=1}^{\infty} \delta(1-\delta)^{t-1}\beta^t h = \frac{h\beta\delta}{1 - \beta + \beta\delta}$$

The above reasoning assumes that $\tau$ reaches zero at each stage of the process, a fact which cannot be assumed *a priori*, since it depends crucially on the particular (approximate) belief states that emerge during the monitoring of the process. Unfortunately, strong *a priori* bounds, as a simple function of $\varepsilon$ and $\delta$, are not possible if $\tau > 0$ at more than one stage. The main reason for this is that the conditional plans that one executes generally do not correspond to $\alpha$-vectors that make up the optimal value function. Specifically, when one chooses a $\tau$-optimal vector (for some $0 < \tau \leq 2\varepsilon$) at a specific stage, a (worst-case) error of $\tau$ is introduced *should this be the only stage at which a suboptimal vector is chosen*. If a $\tau$-optimal vector is chosen at some later stage ($\tau > 0$), the corresponding policy is $\tau$-optimal with respect to a vector that is itself only approximately optimal. Unfortunately, after this second "switch" to a suboptimal vector, the error with respect to the original optimal vector cannot be (usefully) bounded using the information at hand.[5]

However, even without these *a priori* guarantees on decision quality, we expect that in practice, the following approximate error bound will work quite well, specifically as a guide to determining appropriate sample complexity, as discussed below:

$$E \leq \frac{2\varepsilon\beta}{1 - \beta} + \frac{2\varepsilon h\beta\delta}{1 - \beta + \beta\delta} \qquad (4)$$

Intuitively, at each stage of the process a $2\varepsilon$-optimal vector will be chosen with high probability. Though we cannot ensure this, in practice we expect that the cumulative error over those stages where *mistakes* are not made can be usefully estimated by the first term. The second term accounts for the possibility of *mistakes*, as in Theorem 4. Here a *mistake* refers to the probability $1-\delta$ event of choosing a vector at a specific stage that is not $2\varepsilon$-optimal.

We also note that a *post hoc* analysis like that described for one-stage analysis can be used to bound error:

**Proposition 5** *Let $t$ be the first stage of the process at which $\tau > 0$, and $t + k$ be the second such stage. Then*

$$E \leq \frac{h\beta\delta}{1 - \beta + \beta\delta} + \beta^{t+1}2\varepsilon + \beta^{t+k+1}h$$

The first term in this bound denotes the error associated with *mistakes*. The second term reflects the $2\varepsilon$ bound on er-ror associated with the first switch to an approximately optimal vector at stage $t$, while the third reflects the second switch. The main weakness in the bound again lies in this last term and its reliance on $h$ to bound error after a second switch. One way in which these bounds can be strengthened is through the use of *switch set analysis*, a technique described in [17]. The set of constraints imposed by the sampling scheme on the true belief state are linear and *a priori* error bounds can be computed by dynamic programming. Details are beyond the scope of this paper.

---

[5]In particular, it is not the case that the error is bounded by $2\tau$ [17].

### 3.5 Dynamic Sample Generation

The analysis above allows us to determine *a priori* the sample complexity required to achieve a certain error with a specified probability. Our objective is ultimately to be reasonably sure we choose the correct (maximizing) $\alpha$-vector at each stage of the process. The method above ensures this by requiring that $V_\alpha^t$ is estimated reasonably precisely for each $\alpha$. The *post hoc* analysis of value separation suggests that *great precision is not needed* if the vectors are widely separated at the true belief state, specifically, if the best vector has value much greater than the second best. Drawing on ideas from the literature on group sequential methods [12] and multiple-comparisons with the best (MCB) [10] that analyze decision making from this perspective, we describe a method that at each stage generates samples dynamically, using a sampling plan whose termination depends on results at earlier stages of the plan. The method is inherently simple: we will take samples in batches until we can select an $\alpha$-vector satisfying certain requirements. Our method recalls the application of MCB results and group sequential methods by Ortiz and Kaelbling to influence diagrams (see [15] for details and further references).

Suppose we are trying to select the maximizing $\alpha$-vector at stage $t$, using belief state $\tilde{b}^t$. The basic structure of our dynamic approach requires that we generate samples from $T(\tilde{b}^t, a^t, o^{t+1})$ in batches, each of some predetermined size. To generate the $j$th batch:

(a) we determine a suitable confidence parameter $\delta_j$

(b) we generate the $j$th batch of $m_j$ samples from $T(\tilde{b}^t, a^t, o^{t+1})$

(c) we compute estimates $\tilde{V}_\alpha^t[j]$ for all vectors $\alpha$ based on the samples in all $j$ batches, corresponding precisions $\varepsilon_\alpha[j]$, and let $\alpha_j^*$ be the vector with greatest value $\tilde{V}_{\alpha_j^*}^t[j]$

(d) we compute threshold $-\tau_j = \tilde{V}_{\alpha_j^*}^t[j] - \varepsilon_{\alpha_j^*}[j] - \max_{\alpha \neq \alpha_j^*}(\tilde{V}_\alpha^t[j] + \varepsilon_\alpha[j])$ and terminate if $\tau_j$ reaches a certain stopping criterion

We now elaborate on this procedure.

We use MCB results to obtain confidence lower bounds (or one-sided confidence intervals) on the difference in true value between that of the vector with largest value estimate with respect to all the samples in the batches so far and the best of the other vectors. Suppose $m_1$ samples are generated in the first batch. Given simultaneous confidence parameter $\delta_1$, we obtain the one-sided bounds $\varepsilon_\alpha[j]$ according to Eq. 1 using $\delta = \delta_1/|\aleph|$ as the individual confidence parameter and $n^t = m_1$ as the number of samples. Defining $\tau_1$ as above, and combining a lower bound for $\alpha_1^*$ with an upper bound for all the others, we have

$$\Pr(V_{\alpha_1^*}^t - \max_{\alpha \neq \alpha_1^*} V_\alpha^t \geq -\tau_1) \geq 1 - \delta_1 \qquad (5)$$

If $\tau = \tau_1$ is nonpositive, $\alpha_1^*$ is the optimal vector with probability at least $1 - \delta_1$. In general, if we stop immediately after processing the first batch and select $\alpha_1^*$, the error incurred will be at most $\max(0, \tau) \leq 2\varepsilon_1 = 2 \max_\alpha \varepsilon_\alpha[1]$.

If we are unsatisfied with the precision $\tau$ achieved, we generate a second batch of $m_2$ samples, and propose that

$$\Pr(V_{\alpha_2^*}^t - \max_{\alpha \neq \alpha_2^*} V_\alpha^t \geq -\tau_2) \geq 1 - \delta_2$$

This bound holds if we insist *beforehand* that we will generate the second batch; but it ignores that fact that we generate this batch only after realizing our stopping condition was not satisfied using the first batch. This dependence on the bound resulting from the first batch—since these bounds are random variables, this means we do not know *a priori* whether we will generate a second batch—requires that we correct for *multiple looks* at the data. We do this by insisting that both bounds hold jointly, conjoining the bounds obtained after two batches using the Bonferroni inequality and letting $\tau = \min_{\{j | j \leq 2, \alpha_j^* = \alpha_2^*\}} \tau_j$:

$$\Pr(V_{\alpha_2^*}^t - \max_{\alpha \neq \alpha_2^*} V_\alpha^t \geq -\tau) \geq 1 - (\delta_1 + \delta_2)$$

Hence, if we stop after processing at most 2 batches, then our error in selecting $\alpha_2^*$ will be at most $\max(0, \tau)$ with probability at least $1 - (\delta_1 + \delta_2)$. Applying this argument up to $k$ batches, we obtain

$$\Pr(V_{\alpha_k^*}^t - \max_{\alpha \neq \alpha_k^*} V_\alpha^t \geq -\tau) \geq 1 - \sum_{j=1}^k \delta_j$$

where $\tau = \min_{\{j | j \leq k, \alpha_j^* = \alpha_k^*\}} \tau_j$.

The method *as described above* will stop at the first batch $l$ such that $\tau_j \leq 0$: at this point we are assured of selecting the optimal vector with high probability. If we insist that we force $\tau$ to zero, the number of batches $k$ cannot be bounded; thus, we must set the sequence of confidence parameters $\delta_j$ such that $\sum_{j=1}^\infty \delta_j \leq \delta$. For example, we might set $\delta_j = \delta/(j(j+1))$ and the individual confidence parameters as $\delta_j/|\aleph|$. If there is separation between the value of the *optimal vector* and the second best, the process will stop after a finite number of batches. Hence, we can continue the process until $\tau = 0$. However, since the error in the individual estimates decreases only proportionally to $\sqrt{\ln j/j}$, termination might take longer than we wish, depending on the amount of separation and the vector-value variance. This problem is exacerbated by our use of loose ranges in the computation of the precisions $\varepsilon_\alpha[j]$.

If we impose a limit $B$ on the number of batches, and want to make sure that our assessment of $\tau$ holds with probability at least $1 - \delta$, we need to set the sequence of confidence parameters $\delta_j$ such that $\sum_{j=1}^B \delta_j \leq \delta$. The easiest way to accomplish our global confidence requirements is to set $\delta_j = \delta/B$. Furthermore, if we want to be sure that the method selects a vector with true value that is no less than $2\epsilon$ from the optimal with the same confidence, then one alternative is to set the number of samples $m_j$ in each batch $j$ to the $\lceil (\max_\alpha R_\alpha^2/(2B\epsilon^2)) \ln(B|\aleph|/\delta) \rceil$. If we do not impose specific requirements then the setting of $m_j$ is arbitrary, but needs to be fixed in advance. This is because for our analysis to hold, $m_j$ cannot depend on the outcomes from the samples themselves. Although arbitrary, in general, the setting of $m_j$ should take into consideration a trade-off between reducing the expected total number of samples before the method stop versus reducing the variation on the total number of samples.

In general, we expect this method to be effective when there is sufficiently large gap between the best vector and the rest, and/or the ranges in vector values are sufficiently small relative to the value separation and the error tolerance. By using loose upper bounds on the variances and accuracy parameters, the theoretical bounds can become very loose, and hence do not reflect the potential gains we expect. The version presented in this paper is very simple. Many variations on the same idea are possible to try to bring the theoretical bounds more in accordance with our belief about the expected behavior of the method (for instance, using information about range of differences in value between vector pairs, allocating some samples to estimate variance, etc.), but this is beyond the scope of this paper.

As before, unless we push the error tolerance $\tau$ to zero at each stage of the monitoring process, we cannot obtain tight bounds on error after that point. However, we can assert:

**Theorem 6** *Suppose beliefs are monitored according to the dynamic procedure described above using global confidence parameter $\delta$. Furthermore, suppose that $\tau = 0$ at each stage. Then error $E$ satisfies*

$$E \le \frac{h\beta\delta}{1 - \beta + \beta\delta}$$

However, as noted above, the computational demands of insisting that $\tau = 0$ can be severe if the belief state at some time $t$ is such that little separation exists between the best vector and the second-best (that is, if $\hat{b}^t$ lies close to a "edge" of the value function, where two optimal $\alpha$-vectors intersect). If $\tau \le 0$ at all stages up to time $t$, then the bound described in Proposition 5 holds for this dynamic scheme.

## 4 Empirical Evaluation

Three test problems were used to carry out experiments testing the efficacy of our sampling procedures (we refer to [16] for the full specification of those problems; see also [18] for a summary). Each of the three problems was solved using Hansen and Feng's [9] ADD implementation of incremental pruning (IP) to produce a set $\aleph$ of $\alpha$-vectors using a compact ADD representation.

In the following experiments, we report on the use of sampling for approximate belief state monitoring on three test problems. The goal of the experiments are twofold: to evaluate (i) the impact on decision quality induced by sampling techniques and (ii) the sample complexity necessary to guarantee some level of decision quality. Note that the experiments do not evaluate the running time of sampling methods since that is not the focus of this paper and the efficiency gains of such methods have already been clearly demonstrated [11, 21]. In theory, exact monitoring has time complexity on the order of $O(|S|^2)$ whereas sampling has a time complexity in the order of $O(m \log |S|)$ ($m$ is the number of samples). Thus, a sampling strategy provides time savings when $m < |S|^2/\log |S|$. The reader should also be warned that the scope of the empirical evaluation was limited to test problems for which a set of $\alpha$-vectors corresponding to an optimal value function can be computed. Hence, as shown in Table 1, $|S|$ and $|\aleph|$ are fairly small, and consequently the following experiments should be considered preliminary.

| Problem | State Space Size | Size of $\aleph$ | |
|---|---|---|---|
| | | maximum | average |
| Coffee | 32 | 102 | 56 |
| Widget | 32 | 205 | 121 |
| Pavement | 128 | 39 | 16 |

Table 1: Statistics for the three test problems. The maximum and average size of $\aleph$ are taken over a 15-stage process.

The first experiment compares the expected loss incurred by sampling methods to that of a random monitoring approach. More precisely, 5000 initial belief states are picked uniformly at random and for each initial belief state, the optimal expected total reward is compared to the cumulative rewards earned by an agent that approximately monitors its belief state over 15 stages. The difference between the optimal expected total return and the actual return is the loss due to approximate monitoring. Table 2 shows the average loss due to a *single* approximation at the first stage (assuming exact monitoring for the remaining 14 stages), whereas Table 3 shows the average *cumulative* loss due to approximate monitoring at each of the 15 stages. When doing random monitoring, the agent picks a belief state at random (uniformly) and executes the optimal action for this random belief state. This random method can be viewed as a naive strategy that any other approximation method should be able to beat. The sampling methods implemented are basic particle filtering (with partial evidence integration) where a fixed number of particles (20, 40, 80 or 160) are sampled for each approximate belief state. The column "worst" reports the worst possible expected loss that can be achieved by consistently choosing the worst actions.[6] The worst expected loss is included to give some idea of the scale of potential losses due to approximate monitoring.

As expected, the experiments show a gradual decrease in average expected loss as the number of samples increases. When compared to the random strategy (and considering the range of values obtainable across the set of possible behaviors), sampling methods perform quite well. In Table 2, the first row of each problem indicates the actual error incurred and the second row indicates the upper bound $2\epsilon$ predicted by the theory (for $\delta = 0.1$). This bound is loose when compared to the actual error due to the worst-case nature of the analysis. The bounds may still provide some guidance regarding the amount of sampling desired to reduce the average expected loss to some suitable level (assuming a more or less constant ratio between the bounds and the actual error).

---
[6]This worst strategy can be computed by minimizing (instead of maximizing) the expected total reward while solving the POMDP.

| Prob | Average Single Error | | | | | |
|---|---|---|---|---|---|---|
| | Rand | Sampling | | | | Worst |
| | | 20 | 40 | 80 | 160 | |
| Coff | 0.261 | 0.008 | 0.005 | 0.003 | 0.002 | 1.263 |
| $2\epsilon$ | | 1.796 | 1.270 | 0.898 | 0.635 | |
| Widg | 0.101 | 0.034 | 0.021 | 0.012 | 0.007 | 1.099 |
| $2\epsilon$ | | 0.315 | 0.223 | 0.158 | 0.111 | |
| Pav | 0.201 | 0.030 | 0.020 | 0.013 | 0.009 | 1.968 |
| $2\epsilon$ | | 1.266 | 0.895 | 0.633 | 0.448 | |

Table 2: Comparison of the average error due to a single approximation at the first stage of a 15-stage process (exact monitoring being performed for the remaining 14 stages).

| Prob | Average Cumulative Error | | | | | |
|---|---|---|---|---|---|---|
| | Rand | Sampling | | | | Worst |
| | | 20 | 40 | 80 | 160 | |
| Coff | 1.653 | 0.100 | 0.043 | 0.018 | 0.017 | 8.014 |
| Widg | 0.109 | 0.098 | 0.069 | 0.045 | 0.022 | 5.778 |
| Pav | 2.319 | 0.124 | 0.072 | 0.045 | 0.024 | 34.24 |

Table 3: Comparison of the average cumulative error due to approximate monitoring at each stage of a 15-stage process.

In a second experiment, we evaluate the benefits of dynamically determining the amount of sampling. For given $\delta$ and $\epsilon$, we evaluate the total number of samples necessary to guarantee that the one-stage sampling error is bounded by $2\epsilon$ with confidence $1-\delta$. Table 4 shows how this total number of samples varies as we increase the maximum number of batches. Once again, 5000 random initial belief states are chosen and the average number of samples required to decrease $\tau$ below $2\epsilon$ is reported. The column for 1 batch corresponds to the standard non-dynamic sampling procedure. Table 4 reveals that for the widget and pavement problems, a dynamic sampling procedure can reduce the sampling complexity quite dramatically for a well-chosen maximum number of batches. Unfortunately, the dynamic approach does not appear to have offered any savings in the coffee problem. Further investigation is necessary to assess the optimal (maximum) number of batches in general.

In a related paper [18], Poupart and Boutilier also tackle the belief state monitoring problem, but using a vector space method that exploits conditional independence. The idea is to repeatedly approximate belief states using projections as initially proposed by Boyen and Koller [3]. Projection schemes and sampling approaches differ in many aspects including the properties of POMDPs for which they are most suitable. Sampling methods exploit the sparsity of belief distribution whereas projection schemes exploit conditional independence. Given that the coffee, widget and pavement problems are factored POMDPs, the vector space methods tend to perform better than sampling with respect to decision quality. For instance, average losses due to single-stage approximation using the max VS-search method are respectively 0.0013, 0.0082, 0.0014 for the coffee, widget and pavement problems; similarly, the average cumulative losses over 15 stages are respectively 0.0154, 0.0519 and 0.0071. However, the computational overhead associated with sampling is minimal while the overhead associated with choosing good projection schemes is nontrivial. We expect the two approaches can be combined in fruitful ways (as we discuss below).

| Prob | Maximum number of batches | | | | | | | | | |
|---|---|---|---|---|---|---|---|---|---|---|
| | 1 | 2 | 3 | 4 | 5 | 6 | 7 | 8 | 9 | 10 |
| Coff | 258 | 266 | 278 | 277 | 250 | 256 | 267 | 248 | 254 | 265 |
| Widg | 139 | 107 | 93 | 84 | 82 | 86 | 80 | 78 | 78 | 80 |
| Pav | 106 | 64 | 62 | 52 | 66 | 62 | 60 | 55 | 58 | 59 |

Table 4: Comparison of the average number of samples required for adaptive sampling at the first stage of a 15-stage process ($\delta = 0.1$ and $\epsilon = 2$ for coffee and pavement, $\delta = 0.1$ and $\epsilon = 0.5$ for widget).

## 5 Concluding Remarks

Our value-directed sampling technique can be seen as applying methods from the MCB and group sequential sampling fields to the problem of particle filtering for POMDPs. We are able to derive (worst-case) error bounds on such an approach, and use these bounds to suggest methods to direct sampling in such a way as to choose optimal actions rather than (necessarily) accurately estimate their values. Our initial empirical results are encouraging, though clearly much more substantial testing is needed, a task in which we are currently engaged.

This research can be extended in a number of ways in a number of very interesting ways. One important challenge is to provide a stronger analysis of error when the precision parameter $\tau > 0$. One strategy to circumvent this difficulty builds on the idea of constructing the set of *alternative* conditional plans that may be executed when $\tau > 0$ [17]. Another challenge is to provide an analysis in the absence of partial EI (which locally removes bias): one idea is to use information from the DBN parameters to compute *a priori* error bounds; another is to use absolute approximation bounds similar to those used in this paper or optimal relative approximation methods to obtain *a posteriori* bounds on the error tolerance $\tau$.

We are very interested in adapting these techniques to other value function representations (e.g., grid-based value functions) and providing an error analysis of this method when the value function is itself an approximation of the true value function. Finally, previous work using value-directed projection schemes [3, 17] has been used successfully to ex-ploit the conditional independence present in certain factored POMDPs to speed up belief monitoring. The sampling approach described in this work does not exploit this type of structure; however, one could sample the variables defining the factored state space in a "stratified" fashion, or apply Rao-Blackwellisation methods [6, 7].

**Acknowledgements:** Boutilier and Poupart were supported by the Natural Sciences and Engineering Research Council and the Institute for Robotics and Intelligent Systems. Ortiz was supported by NSF IGERT award SBR 9870676.